\documentclass{article}

      \usepackage[preprint]{neurips_2020}

\usepackage[utf8]{inputenc} 
\usepackage[T1]{fontenc}    
\usepackage{hyperref}       
\usepackage{url}            
\usepackage{booktabs}       
\usepackage{amsfonts}       
\usepackage{nicefrac}       
\usepackage{microtype}      
\usepackage{algorithm}
\usepackage{algorithmic}
\usepackage{amsmath}
\usepackage{pythonhighlight}
\usepackage{graphicx}
\usepackage{enumerate}
\usepackage{indentfirst} 
\usepackage{float}
\title{Masked Face Image Classification with Sparse Representation based on Majority Voting Mechanism\thanks{This paper serves as the final project report for the course "Representation Learning" by Professor Zhang Lei in SiChuan University.} }

\author{
  Wang Han\\
  Department of Computer Science\\
  SiChuan University\\
  \texttt{2018141461062@stu.scu.edu.cn} \\
}

\begin{document}

\maketitle

\begin{abstract}
Sparse approximation is the problem to find the sparsest linear combination for a signal from a redundant dictionary, which is widely applied in signal processing and compressed sensing. 
In this project, I manage to implement the Orthogonal Matching Pursuit (OMP) algorithm and Sparse Representation-based Classification (SRC) algorithm, then use them to finish the task of masked image classification with majority voting. 
Here the experiment was token on the AR data-set, and the result shows the superiority of OMP algorithm combined with SRC algorithm over masked face image classification with an accuracy of 98.4\%. 

\end{abstract}

\section{Introduction}
In the field of signal processing, it is the usual case to represent signal with a linear combination of set of m unit vectors. When those vectors are orthonormal, this approximation can be regarded as projection of signal onto the subspace spanned by elements of this m orthonormal basis. However, when those vectors are linear dependent, less than m vectors is needed since $m-r$ non-independent elements can be represent by the other r linear independent vectors.
It is easy to imagine that if m orthonormal basis can perfectly approximate a signal, surely m + 1 should not be worse. Though orthogonal basis is efficient in most case, a dictionary with linear dependent vectors can do even better. That’s why sparse approximation is of great significance in signal processing.Under
the assumption of a linear dependent set, or in other words, redundant dictionaries, finding the sparsest representation is called the sparse approximation problem.
One of the the biggest challenge for sparse approximation is the selection of atoms for the sparsest representation, which make sparse approximation a NP-hard problem.

With the superiority of OMP, a question rises:"Will there be satisfying result if using OMP algorithm to solve the problem of masked face image classification"? 
Here an experiment was carried out to test the performance of this idea. 
Firstly, each image from the training dataset was downsampled and divided into multiple grids as well as the testing images to ensure I.I.D. 
Then the training samples were concatenated to generate the dictionary. 
Secondly, OMP algorithm was implemented with this dictionary and each testing image grid as input to get sparse representation for the input grid. 
Finally, majority voting was done on the grids to get the prediction of each testing image. 

In the experiment, I tried different grid sizes, down-sample resolutions and sparse representation shapes.
The statistics show that best result can be achieved with 11x11 grid size, 55x66 down-sample resolution and 1500x1 sparse representation shape. The best result comes with an accuracy of 98.4\% over testing dataset.

\section{Related work}
\subsection{Robust Face Recognition via Sparse Representation}
In the research by John Wright,et al[5], the problem is studied that automatically recognizing human faces from frontal views with varying expression and illumination, as well as occlusion and disguise. 
We cast the recognition problem as one of classifying among multiple linear regression models and argue that new theory from sparse signal representation offers the key to addressing this problem. 
Based on a sparse representation computed by l1-minimization, we propose a general classification algorithm for (image-based) object recognition. 
This new framework provides new insights into two crucial issues in face recognition: feature extraction and robustness to occlusion. 
For feature extraction, we show that if sparsity in the recognition problem is properly harnessed, the choice of features is no longer critical.
What is critical, however, is whether the number of features is sufficiently large and whether the sparse representation is correctly
computed. 

Unconventional features such as downsampled images and random projections perform just as well as conventional features such as Eigenfaces and Laplacianfaces, as long as the dimension of the feature space surpasses certain threshold, predicted by the theory of sparse representation. 
This framework can handle errors due to occlusion and corruption uniformly by exploiting the fact that these errors are often sparse with respect to the standard (pixel) basis.

The theory of sparse representation helps predict how much occlusion the recognition algorithm can handle and how to choose the training images to maximize robustness to occlusion. 

\subsection{Theoretical Analysis of the Application of Majority Voting to Pattern Recognition}
In the research by Louisa Lam and Ching Y. Suen[6], it's assumed that n classifiers or experts are used, and that for each input sample, each
expert produces a unique decision regarding the identity of the sample. 
This identity could be one of the allowable classes, or a rejection when no such identity is considered possible. 
In combining the decisions of the n experts, the sample is assigned the class for which there is a consensus, or when more than half of the experts are agreed on the identity. 
Otherwise the sample is rejected.
While each classifier has the possibilities of being correct, wrong, or neutral, the combined (correct) recognition rate is really the probability of the consensus being correct, assuming each vote to have only 2 values correct or not.

Due to the nature of consensus, the combined decision is wrong only when a majority of the votes are wrong and they make the same mistake. This is a strength of this combination method - due to the large number of possible mistakes, the majority would not often make the same one. 

As a result of this need for consensus, we can only calculate the probability of the coflsemus committing a particular error from the individual probabilities of committing the same error.

\section{Methodology}

\subsection{Sparse Representation}

Consider a linear system of equations x = D$\alpha$, where D is an underdetermined $m * p$ matrix (m<p) and $x \in R^{m}$, $a \in R^{p}$. The matrix D (typically assumed to be full-rank) is referred to as the dictionary, and x is a signal of interest. The core sparse representation problem is defined as the quest for the sparsest possible representation $\alpha$ satisfying $x=D\alpha$. Due to the underdetermined nature of $D$, the linear system admits in general infinitely many possible solutions, and among these we seek the one with the fewset non-zeros. Put formally, we solve the problem

\begin{equation}
    min_{a \in R^{P}}||a||_{0} \ \ subject \ to \ x=D\alpha
\end{equation}

where $||a||_{0} = a_{i}, i=0,i=1,...,p$ is the $l_{0}$ pseudo-norm, which counts the number of non-zero components of $\alpha$ . This problem is known to be NP-Hard with a reduction to NP-complete subset selection problems in $combinatorial$ optimization.

Sparsity of $\alpha$  implies that only a few ({$\displaystyle k\ll m<p$}) components in it are non-zero. The underlying motivation for such a sparse decomposition is the desire to provide the simplest possible explanation of $x$ as a linear combination of as few as possible columns from D, also referred to as atoms. As such, the signal $x$ can be viewed as a molecule composed of a few fundamental elements taken from $D$.

\subsection{Orthogonal Match Pursuit }
In this section, we give a detailed description of the orthogonal matching pursuit(OMP) algorithm. 
For any subset $S \subset \Omega$, denote by A(S) a sub-matrix of measurement matrix A consisting of the columns $\psi_{i}$ with $i \in S$. 
X(S) is the sub-vector corresponding to A(S).
Thus, the OMP algorithm can be stated as follows.

\begin{algorithm} 
\caption{The Orthogonal Matching Pursuit Algorithm}
\label{alg1} 
\begin{algorithmic}[1] 
\REQUIRE Measurement \ $y$, \ measurement \ matrix $A = [\psi_{1}, \psi_{2}, ..., \psi_{N} ]$  
\ENSURE Recovered signal x, \ indices for corresponding atoms indexes S 
\WHILE{ $stop-rule == FALSE$ } 
\STATE Find $i^{*}_{m}$ $\in$ $\Omega$, that $i^{*}_{n}$ = $argmax_{i}|<r_{m}, \psi_{i}>|$
\STATE $S \gets S \cup i^{*}_{m}$
\STATE $x \gets (A(S)^{T}A(S))^{-1}A(S)^{T}y$
\STATE $r_{m+1} \gets r_{m} - Ax $
\STATE $m \gets m+1$
\ENDWHILE 
\end{algorithmic} 
\end{algorithm}

The OMP is also a step-wise forward selection algorithm and is easy to implement. Followed by the same atom selection criteria as Match Pursuit(MP) method, OMP differ from MP in that, at each step, OMP computes the least square solution for y = A(S)x that $x = (A(S)^{T}A(S))^{-1}A(S)^{T}y$. This least-squares minimization aims to obtain the best approximation over the atoms that have already been chosen. 
A superiority of OMP is that it never selects the same atom twice because the residual is orthogonal to the atoms that have already been chosen.\\
Another key component of OMP is the stopping rule which depends on the noise structure. 
In the noiseless case the natural stopping rule is $r_{m}$ = 0. In this project, we shall consider several different noise structures.
To be more specific, two types of noise condition are considered. One is that when the sparsity s = |S| is known, the iteration stops when $n \ge s$. 
Another situation stands for unknown sparsity but known norm-bound. 
Under the second condition, the searching for S stops when $||r_{m}||_{2} \leq t$ where $r_{m} = y-Ax$. 
In addition, we would mainly consider the important case of Gaussian noise where $n \sim N(0, \sigma^{2} )$.

\subsection{Sparse Representation-based Classification }

SRC formulated in [5] belongs to the reconstructive classification approach which aims at tackling the classification problem on data with corruption (i.e. noise, missing data and outliers). The details can be shown as below.

\begin{algorithm}[H]
\caption{ The Sparse Representation-based Classification Algorithm}
\label{alg1} 
\begin{algorithmic}[1] 
\STATE \textbf{Input}: a matrix of training samples $A = {A_1,A_2,..., A_k} \subset R^{m x n}$ for k classes, a test sample $y \in R^{m}$
\STATE Normalize the columns of A to unit l2-norm
\STATE Solve the l1-minimal problem \\ 
\begin{center}
    $x_{1} = argmin_{x}||x||_{1}, \ s.t. \ Ax=y $
\end{center}
\STATE Compute the residuals 
\begin{center}
    $r_{i}(y)=||y-A\phi_{i}(x_{1})||_{2}$ , for $i = 1,...,c$
\end{center}
where $\phi_{i}:R^{n} \to R^{n}$ is the characteristic function that selects the coefficients associated with the $i$th class 
\STATE \textbf{Output}: identity(y) = $argmin_{i}r_{i}(y)$
\end{algorithmic}
\end{algorithm}

The second step which is used to calculate the sparse decomposition is the core of SRC algorithm. Theoretically, suppose that A can offer an over-complete basis, and then to find the sparse representation, we need to solve the following $l_{0}$-norm minimization problem:

\begin{center}
\begin{equation}
    x_{1} = argmin_{x}||x||_{0}, \ s.t. \ Ax=y
\end{equation}
\end{center}

where $||x||_{0}$ is $l_{0}$-norm which is equivalent to the number of non-zero components in the vector x. Notice that the linear system in Eq.x is under-determined since m << n. Finding the exact solution to Eq.x is NP-hard due to its nature of combinational optimization. An approximate solution is obtained by replacing the $l_{0}$ norm in Eq.x with the $l_{1}$ norm as in Eq.x. It can be proved that the solution of Eq.x is equivalent to the solution of Eq.x  if a certain condition on the sparsity is satisfied, i.e. the solution is sparse enough.

In theory, the computational complexity to obtain the sparse representation is about $O(t^{2}n)$ per test sample, where $t$ is the number of nonzero entries in reconstruction coefficients and $n$ is the number of training samples. In practice,however, the obtained solution is far more from sparse vector due to many very small non-zero reconstruction coefficients.

As a result, the computation complexity to find the ’sparse’ solution tends to about $O(n^{3})$ per test sample due to $t$ tends to $n$.

\section{Experiment}

\subsection{AR dataset}

The AR dataset includes frontal face images of 126 individuals, while each individual has 14 unobstructed images. Therefore, there are 126 classes of face images in the AR dataset. 

For each image, 'm' stands for male sample, while 'w' stands for female sample, the first number in its filename stands for the class it belong to. The last number is its id number in the class.

The training dataset is consisted of images whose id number is in the range 1-7 or 14-20, while the rest images make up the testing dataset.

\subsection{Experiment setup}
The whole process can be summarized as below: 

\begin{enumerate}[i.]

\item Check the image shape of the AR dataset, and decide the grid size, then take the unmasked face images as training data, while masked face images as testing data.

\item Downsample each image of in the training dataset and testing dataset, then reshape them to be 1D vector. \ After that, Normalize all of them with l2-normalization.

\item Concatenate all the vectors from training dataset along raw axis to get dictionary, $A=[A_{1},A{2},A_{3},...,A_{k}]$. 

\item Processing in step 2 was also done on testing data, then use OMP algorithm to get sparse representation for each grid of each testing image.

\item For each image, carry out SRC algorithm on multiple grids, then get the final prediction by majority voting. 

\item Count the classification results and calculate the class prediction accuracy and global prediction accuracy.

\end{enumerate}

\subsection{Experiment result and analysis}

With different grid sizes, the experiment was done under three situations respectively as shown in Figure 1:

\begin{enumerate}[i.]
    \item Grid size is 5x5
    \item Grid size is 7x7
    \item Grid size is 11x11
\end{enumerate}

\begin{figure}[htbp]
\centering
\includegraphics[scale=0.5]{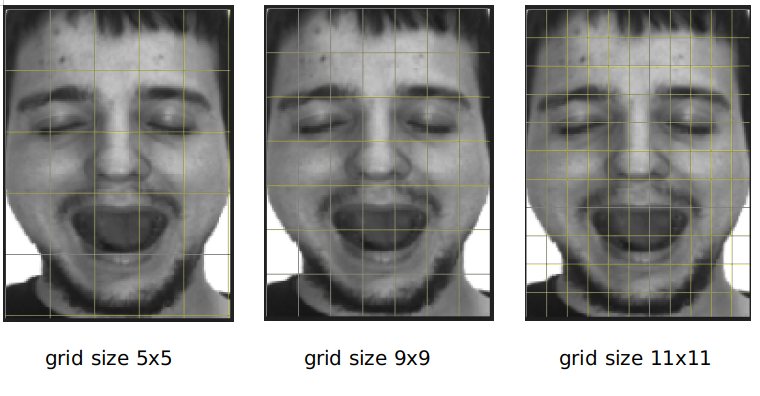}
\caption{Grid Clip Demo}
\end{figure}

The evaluation was done on testing dataset. Figure 2 shows the relation between right predicted image number and predicted image number under three situations. Figure 3 shows the class accuracy of 100 persons under three situations.

\begin{figure}[!hbt]
    \centering{\includegraphics[width=16cm]{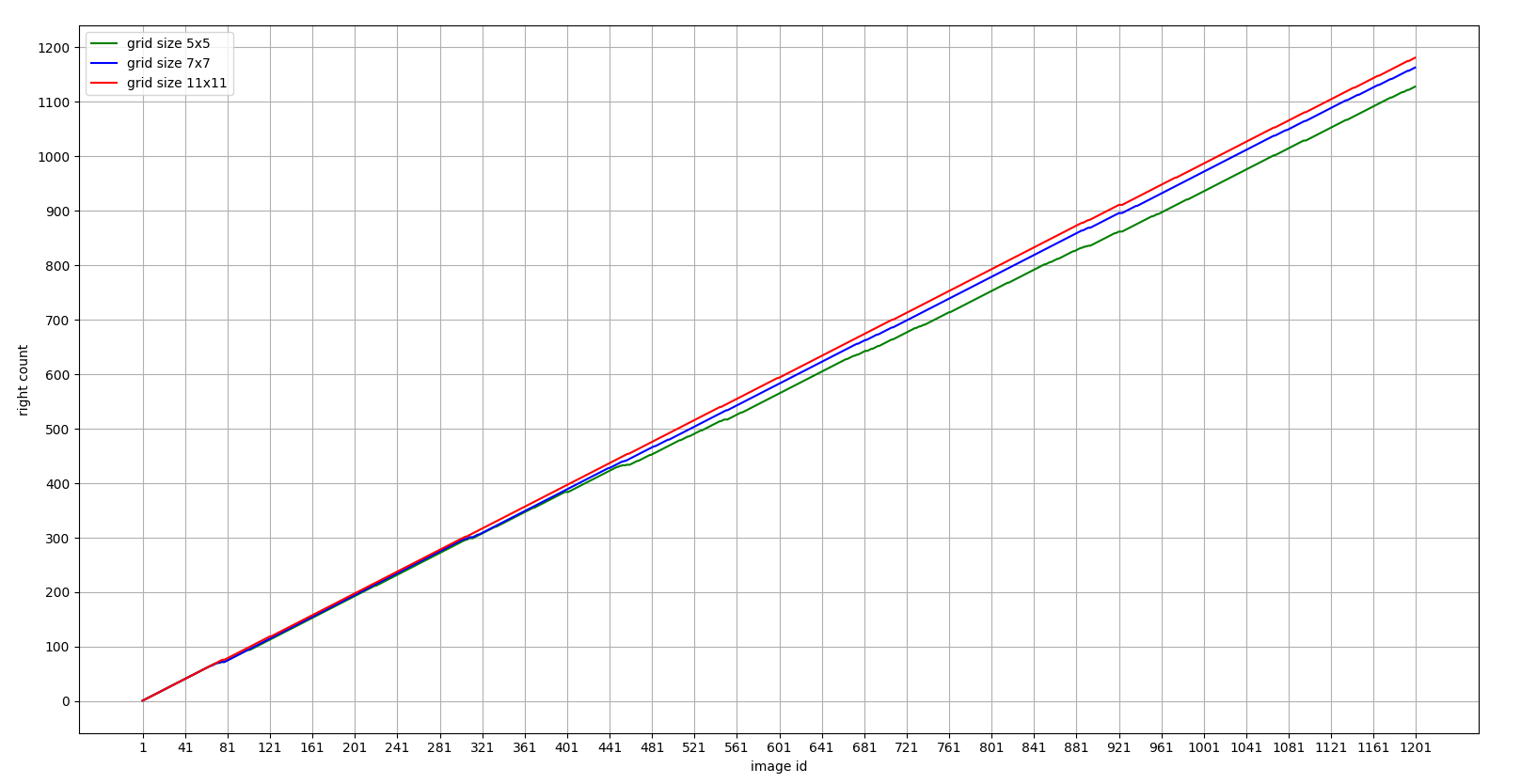}}
    \caption{Right prediction count under different grid sizes}
    \label{fig:my_label1}
\end{figure}

According to the results, we can find that more bigger grid size, more higher accuracy. 

After that, we can get the list of global accuracies under three situations:

\begin{enumerate}[i.]
    \item grid size 5x5: 94.10\% 
    \item Grid size is 7x7: 96.77\% 
    \item Grid size is 11x11: 98.41\% 
\end{enumerate}

As above, the best result was acheived under grid size 11x11 with an accuracy of 98.4\%.

Figure 4 is an accurate statistic of class accuracy under grid size 11x11.

\section{Conclusion}
In this paper, I try to implement sparse representation-based classification algorithm and use it to finish the task of masked face image classification with majority voting. 
By setting grid size to be 11x11, The prediction accuracy over testing dataset has been improved to 98.4\%, which can be considered as a success on this problem.

The drawback is the slow speed of my algorithm, which appears to be 4 seconds for one single 50x60 image on a laptop with an 8-core processor. This can be a terrible bottleneck when the dataset appears to be huge. The future work should consider parallel computation to boost the speed, such as CUDA.  

Another future work I'm still thinking about is add weights to the grid majority voting. This trick may bring amazing result when the task is generalized to large scale and muilti-channel image dataset.

\section*{References}
\medskip
\small

[1] T Tony Cai\ \& Lie Wang\ (2011) Orthogonal matching pursuit for sparse signal recovery with noise. Institute of Electrical and Electronics Engineers.

[2]  Anna C. Gilbert\ {\it TAn introduction to sparse approximation}. Department of Mathematics University of Michigan

[3] Stéphane G Mallat\ \& Zhifeng Zhang.\ Matching pursuits with time-frequency dictionaries. IEEE Transactions on signal processing, 41(12):3397–3415, 1993.

[4] Joel A Tropp. \ Greed is good: Algorithmic results for sparse approximation. {\it IEEE Transactions on Information theory}, 50(10):2231–2242, 2004.

[5] J. Wright\ \&A. Yang\ \&A. Ganesh\ \&S. Sastry\ \&Y. Ma. \ “Robust face recognition via sparse representation”, \ IEEE Trans. on PAMI, vol. 31, no. 2, pp. 210–227, Feb 2009.

[6] Louisa\ Lam,\ \  Ching\ Y. Suen. \ \ "A Theoretical Analysis of the Application of Majority Voting to Pattern Recognition", \ \ 
Centre for Pattern Recognition and Machine Intelligence, 2002

\section*{Appendix}

\subsection*{Code}
\begin{python}
from collections import Counter
from PIL import Image
import numpy as np
import os
import cv2

train_idxs = list(range(1, 8)) + list(range(14, 21))
test_idxs  = list(range(8, 14)) + list(range(21, 27))

def l2_norm(x, units):
    '''
    x       data to process
    units   numbers of unit in x
    '''
    v = 0
    for idx in range(units):
        v += x[idx]**2
    v = np.sqrt(v)
    return x/v

def OMP(y, A, x_size, sp_size):
    # y=Ax
    # A [sp_size, x_size]
    # y [x_size] real signal
    # X [sp_size] 
    # r residual
    Ak = np.zeros([x_size, 0])
    r = y[:, np.newaxis]
    cols = []
    for _ in range(x_size):
        proj = np.abs(np.dot(A.T, r))  # [sp_size]
        best_atom_index = np.argmax(proj)
        cols.append(best_atom_index)
        Ak = np.c_[Ak, A[:, best_atom_index]]
        xk = np.linalg.pinv(Ak.T.dot(Ak)).dot(Ak.T).dot(y[:, np.newaxis])
        r = y[:, np.newaxis] - Ak.dot(xk)
    X = np.zeros(shape=(sp_size,), dtype=np.float32)
    X[cols] = xk.flatten()
    return X

def getDict(PATH, W, H, x_n, y_n, sp_size):
    files = os.listdir(PATH)
    img_shape = (W,H)
    grid_w = img_shape[0]//x_n 
    grid_h = img_shape[1]//y_n

    A = np.zeros(shape=(x_n*y_n, grid_w*grid_h, sp_size), dtype=np.float32)
    TEST = np.zeros(shape=(x_n*y_n, grid_w*grid_h, 1400), dtype=np.float32)

    mask_train = {}
    for file_name in files:
        mask_train[file_name[0:5]] = []

    col_train,col_test = 0,0
    for file_name in sorted(files):
        gender, person_idx, img_idx = file_name.split('-')
        img_idx = int(img_idx[:2])
        person_idx = int(person_idx)
        img = Image.open(os.path.join(PATH, file_name))
        img = np.array(img)
        img = cv2.resize(img, img_shape)
        # divided into blocks
        for i in range(x_n):
            for j in range(y_n):
                patch_idx = i*y_n+j
                patch = img[(grid_h*j):(grid_h*(j+1)), (grid_w*i):(grid_w*(i+1))]
                patch = patch.astype(np.float32).reshape(-1, 1).flatten()
                patch = l2_norm(patch, grid_w*grid_h)

                if img_idx in train_idxs:
                    A[patch_idx, :, col_train] = patch
                else:
                    col_test = 50 if gender=='w' else 0
                    col_test+= person_idx
                    col_test = col_test*len(test_idxs)+test_idxs.index(img_idx)
                    TEST[patch_idx, :, col_test] = patch
        if img_idx in train_idxs:
            mask_train[file_name[0:5]].append(col_train)
            col_train += 1
    for key in mask_train.keys():
        mask_array = np.zeros(sp_size, dtype=np.float32)
        mask_array[mask_train[key]] = 1
        mask_train[key] = mask_array
    return grid_w*grid_h, A, TEST, mask_train

def SRC(p_A, X, y, mask_train):
    residuals = {}   # {(id: residual),...}
    for label in sorted(mask_train.keys()):
        projected_X = X * mask_train[label]
        residuals[label] = np.linalg.norm( y[:, np.newaxis] - p_A.dot(projected_X) )
    return min(residuals, key=residuals.get)

if __name__ == "__main__":
    x_n = 11
    y_n = 11
    sp_size = 1500
    sample_length, A, TEST, mask_train = getDict(r"./AR", 55, 66, x_n, y_n, sp_size)
    print("Get dictionary successfully !!!!\n")

    fo = open("log-11-8.txt",'w')
    all_acc = 0
    cnt=0
    for label in sorted(mask_train.keys()):
        # for each person
        # print("------------------------- person_id: {} -------------------------------".format(label))
        right = 0
        for i,item in enumerate(test_idxs):
            # for each sample of this person
            votes = []
            gender, person_idx = label.split('-')
            col_test = 50 if gender=='w' else 0
            col_test+= int(person_idx)
            col_test = col_test*len(test_idxs)+i
            for i in range(x_n):
                for j in range(y_n):
                    patch_idx = i*y_n+j
                    # for each patch of this picture
                    y = TEST[patch_idx, :, col_test]
                    X = OMP(y, A[patch_idx], sample_length, sp_size)   # sparse representation
                    pred = SRC(A[patch_idx], X, y, mask_train)
                    # print("test_idx {}    patch_idx {} :   {}".format(idx, patch_idx, pred))
                    votes.append(pred)
            pred = Counter(votes).most_common(1)[0][0]
            # print(" -------------- img_id:{}  label:{}   pred:{}-------------- ".format(item,label,pred))
            if label == pred:
                right += 1
            print(item, label, pred, file=fo)
            print(label, item, pred)
            cnt+=1
        all_acc+=right
        acc = right/len(test_idxs)
        print("\n label: ",label, " class Accuracy: ", acc, " all acc: ",all_acc/cnt)
    fo.close()
\end{python}

\subsection*{Experiment statistic}
\begin{figure}[!hbt]
    \centering{\includegraphics[width=16cm]{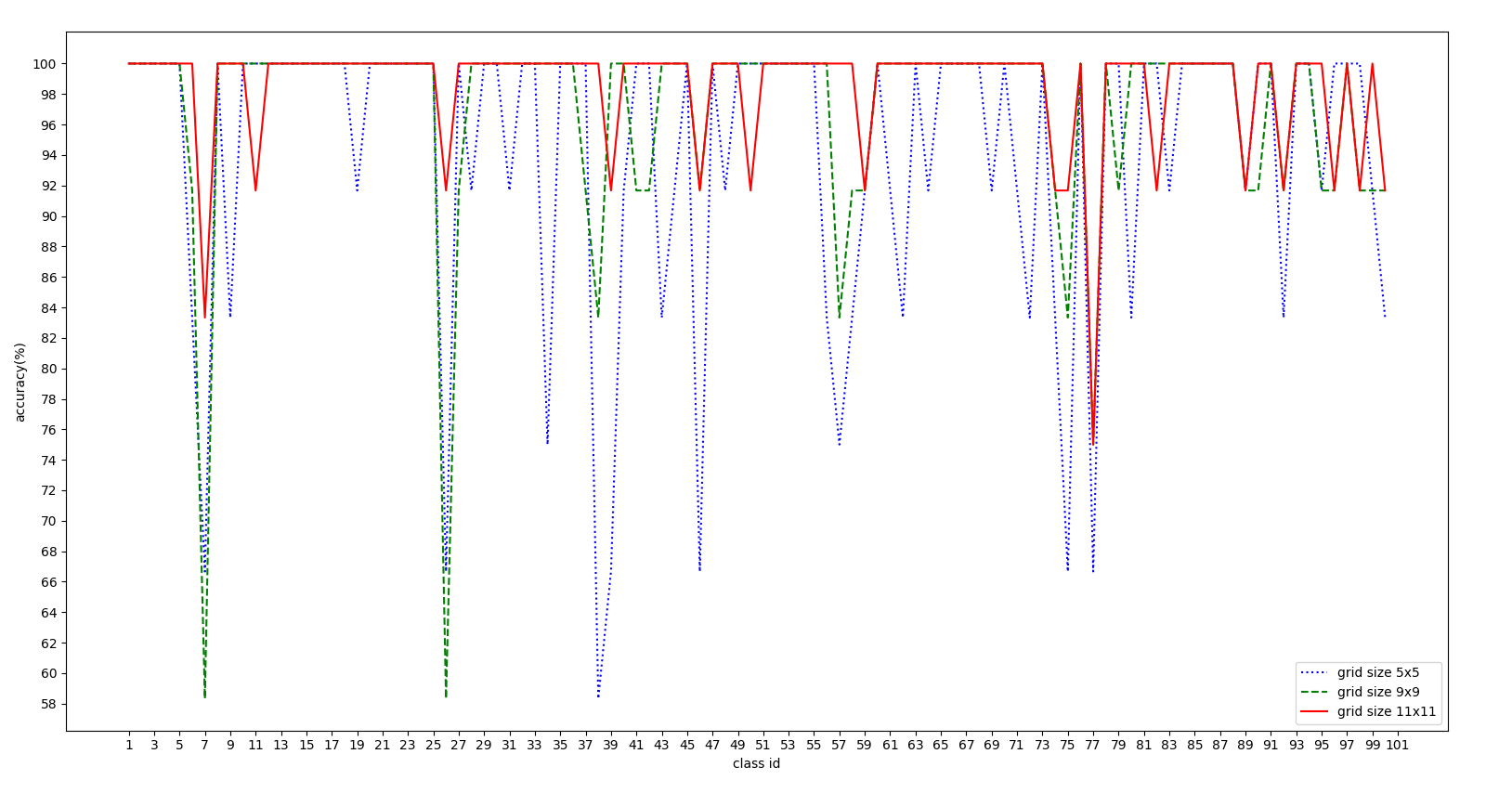}}
    \caption{Class accuracy contrast under different grid sizes}
    \label{fig:my_label2}
\end{figure}

\begin{figure}[!hbt]
    \centering{\includegraphics[width=16cm]{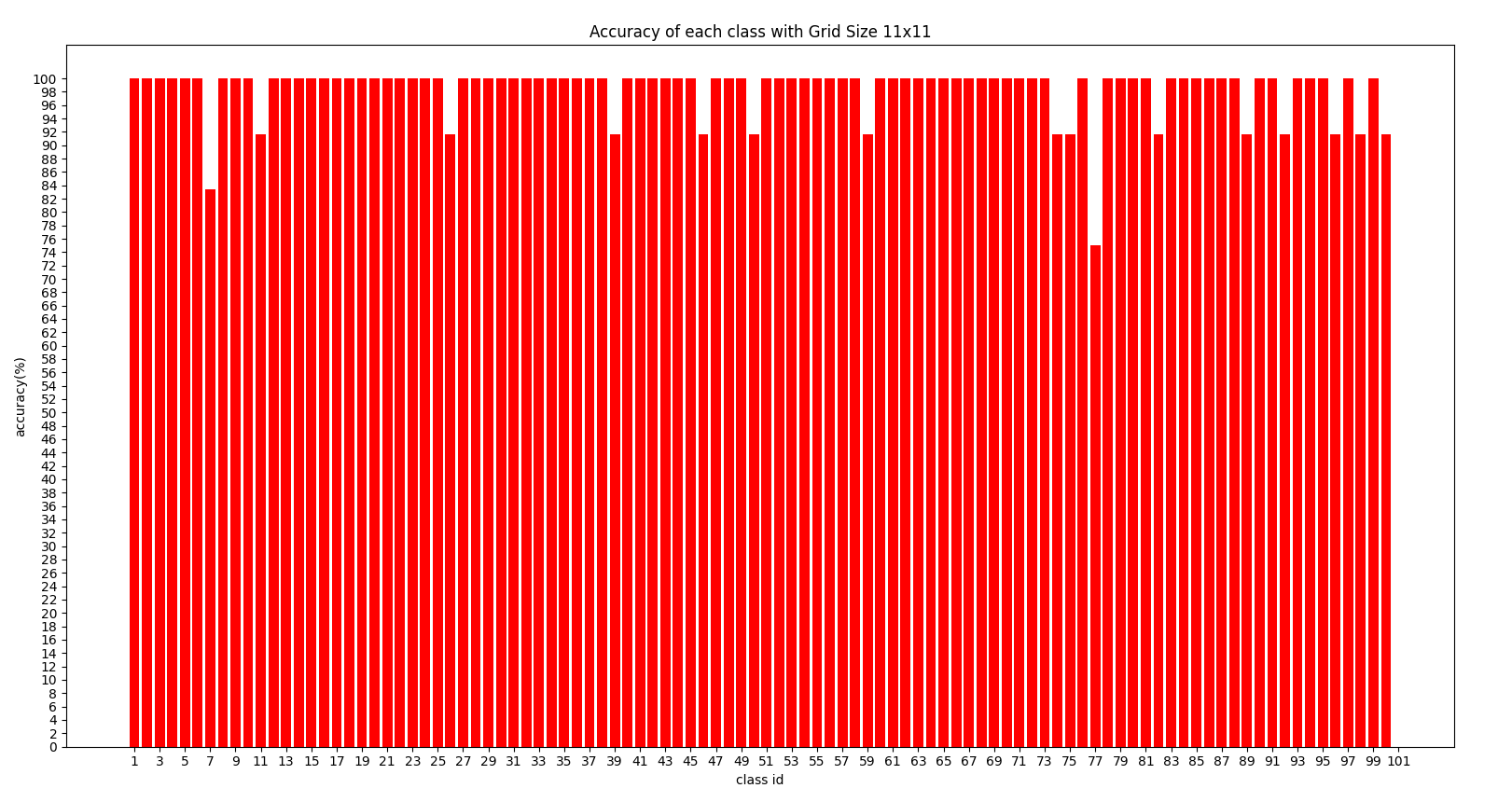}}
    \caption{Class accuracy under grid size 11x11}
    \label{fig:my_label3}
\end{figure}

\end{document}